\documentclass[journal=ancac3,manuscript=article]{achemso}
\usepackage{chemformula} 
\usepackage[T1]{fontenc} 
\usepackage{xcolor}
\usepackage{multirow}
\usepackage{url}
\usepackage{subcaption}
\usepackage[colorlinks, linkcolor=black, citecolor=black, urlcolor=black]{hyperref}
\usepackage{amsmath}
\usepackage{siunitx}     
\usepackage{xspace}      
\usepackage{graphicx}    

\author{Junqi He}
\altaffiliation{These authors contributed equally to this work.}
\author{Yujie Zhang}
\altaffiliation{These authors contributed equally to this work.}
\affiliation[jiliang]{Department of Physics, China Jiliang University, Hangzhou 310018, P. R. China}
\author{Jialu Wang}
\affiliation{Hangzhou Key Laboratory of Quantum Matter, School of Physics, Hangzhou Normal University, Hangzhou 311121, China}
\author{Tao Wang}
\affiliation{ZJU-Hangzhou Global Scientific and Technological Innovation Center, College of Integrated Circuits, Zhejiang University, Hangzhou, 311215, China.}
\author{Pan Zhang}
\affiliation{Department of Physics and Texas Center for Superconductivity, University of Houston, Houston, TX 77204, USA}
\author{Chengjie Cai}
\author{Jinxing Yang}
\affiliation[jiliang]{Department of Physics, China Jiliang University, Hangzhou 310018, P. R. China}
\author{Xiao Lin}
 \affiliation{Key Laboratory for Quantum Materials of Zhejiang Province, Department of Physics, School of Science, Westlake University, Hangzhou 310030, P. R. China}
 \alsoaffiliation{Institute of Natural Sciences, Westlake Institute for Advanced Study, Hangzhou 310024, P. R. China}

\author{Xiaohui Yang}
\email{xiaohuiyang@cjlu.edu.cn}
\affiliation[jiliang]{Department of Physics, China Jiliang University, Hangzhou 310018, P. R. China}
\alsoaffiliation[Zhejiang]{Zhejiang Province Key Laboratory of Quantum Technology and Device, Zhejiang University, Hangzhou 310027, P. R. China}
\alsoaffiliation{Key Laboratory for Quantum Materials of Zhejiang Province, Department of Physics, School of Science, Westlake University, Hangzhou 310030, P. R. China}

\title{Rapid morphology characterization of two-dimensional TMDs and lateral heterostructures based on deep learning}


\begin{document}

\vspace{1ex}
\begin{abstract}
Two-dimensional (2D) materials and heterostructures exhibit unique physical properties, necessitating efficient and accurate characterization methods. Leveraging advancements in artificial intelligence, we introduce a deep learning-based method for efficiently characterizing heterostructures and 2D materials, specifically MoS$_2$-MoSe$_2$ lateral heterostructures and MoS$_2$ flakes with varying shapes and thicknesses. By utilizing YOLO models, we achieve an accuracy rate of over 94.67\% in identifying these materials. Additionally, we explore the application of transfer learning across different materials, which further enhances model performance. This model exhibits robust generalization and anti-interference ability, ensuring reliable results in diverse scenarios. To facilitate practical use, we have developed an application that enables real-time analysis directly from optical microscope images, making the process significantly faster and more cost-effective than traditional methods. This deep learning-driven approach represents a promising tool for the rapid and accurate characterization of 2D materials, opening new avenues for research and development in material science.
\end{abstract}


\section{Introduction}

Two-dimensional (2D) materials have attracted significant attention due to their excellent mechanical, electrical, thermal, and optical properties, making them ideal candidates for next-generation technologies \cite{novoselov2004electric,novoselov2005two,chhowalla2013chemistry,zeng2018exploring,li2020large,doi:10.1126/sciadv.aay8409}. Among these, transition metal dichalcogenides (TMDs) such as MoS$_2$ and MoSe$_2$ stand out for their tunable bandgaps and strong light-matter interactions \cite{zhang2014direct,chhowalla2013chemistry,ganatra2014few,li2015two,eftekhari2017molybdenum,wu2020research}. The properties of 2D materials are highly sensitive to thickness and morphology \cite{jaramillo2007identification,mas20112d,gupta2015recent,tang2020chemical,chhowalla2013chemistry}. For example, single-layer MoS$_2$ has a direct bandgap, making it suitable for optoelectronics, while multilayer MoS$_2$ switches to an indirect bandgap, altering its electronic behavior \cite{chhowalla2013chemistry,ganatra2014few,li2015two}. The performance of 2D materials also depends on shape and size, with large-area flakes showing superior charge transport properties, while nanostructures with edge defects or vacancies exhibit enhanced catalytic activity \cite{lin2018solution,gao2018scalable,jaramillo2007identification,zhang2014dendritic,ye2016defects,tang2020chemical}.

The inherent limitations of individual 2D materials have prompted extensive research into heterostructures, which combine distinct materials to enhance or tailor their properties while preserving the beneficial characteristics of the constituent components\cite{liu2019van,zhang2021more,wang2019van}. These heterostructures are typically classified as lateral or vertical, with lateral heterostructures offering advantages such as reconfigured band structures, reduced interlayer scattering, and superior optoelectronic performance \cite{wang2019van,liu2018interface,schneider2024cvd,behranginia2017direct}. Such attributes make lateral heterostructures particularly promising for next-generation technologies in electronics, optoelectronics, and flexible devices\cite{liu2023controllable,hoang2022two,liu2019borophene}. TMD heterostructures, like MoS$_2$-MoSe$_2$, have been particularly attractive for semiconductor applications due to their remarkable optoelectronic and thermoelectric properties \cite{duan2014lateral,chen2017plane,barik20222d,darboe2022constructing,bellus2018photocarrier,jia2019excellent,xu2024interfacial}.

Due to the diverse physical properties, 2D materials and their heterostructures require highly accurate and efficient characterization techniques. Traditional techniques such as atomic force microscopy (AFM) and Raman spectroscopy are effective but time-consuming\cite{liu2018interface,liang2014first}. In contrast, artificial intelligence, particularly machine learning and deep learning, offers data-driven solutions for efficient material characterization \cite{yang2020automated,li2019rapid,masubuchi2019classifying,ryu2022understanding,lu2024machine,si20232d,choudhary2022recent,https://doi.org/10.1002/adts.202200140}. Models like CNNs, including Mask R-CNN, U-Net, and YOLO, have been used for analyzing optical images of 2D materials, enabling accurate identification of layer thickness, defects, and heterostructures \cite{he2024machine,masubuchi2020deep,dong20213d,shi2022uncovering,yang2024identification,zhu2022artificial}. These methods reduce time and cost compared to traditional measurements and also offer the potential for characterizing physical properties. In the identification of heterostructures, support vector machine (SVM) have been used to distinguish graphene-MoS$_2$ heterostructures, while artificial neural network (ANN) and conditional generative adversarial network (cGAN) have been applied to identify WS$_2$-MoS$_2$ heterostructures\cite{lin2018intelligent,zhu2022artificial,bhawsar2024deep}. Despite their success, current approaches face challenges in portability and require extensive dataset preprocessing, with inference results often needing manual validation.

In this study, we propose a deep learning-based approach for the rapid and accurate characterization of MoS$_2$ flakes and MoS$_2$-MoSe$_2$ lateral heterostructures. Using YOLO models (YOLOv8 and YOLOv11) for object detection and instance segmentation, our method surpasses traditional techniques with faster inference, smaller model size, higher accuracy, and reduced data requirements. We synthesized the materials via chemical vapor deposition (CVD) and characterized them using Raman spectroscopy. An annotated optical image dataset was created, achieving over 96\% recognition accuracy for heterostructures. We also observed that synthesis conditions significantly affect MoS$_2$ morphology, which in turn influences its properties. We developed a shape recognition method using deep learning, achieving over 94\% accuracy in detecting and quantifying nanoscale shapes. Transfer learning was applied to fine-tune a pre-trained heterostructure model on a MoS$_2$ thickness dataset, improving performance by 3\%. The model also demonstrated robust generalization abilities. To streamline the characterization process, we developed a software tool that integrates deep learning models with optical microscope outputs for real-time inference, reducing processing delays. This approach enhances operational efficiency and highlights the practical applicability of deep learning for large-scale material analysis.

\section{Results and discussion}

\begin{figure*}[!thb]
    \centering
    \includegraphics[width=0.75\textwidth]{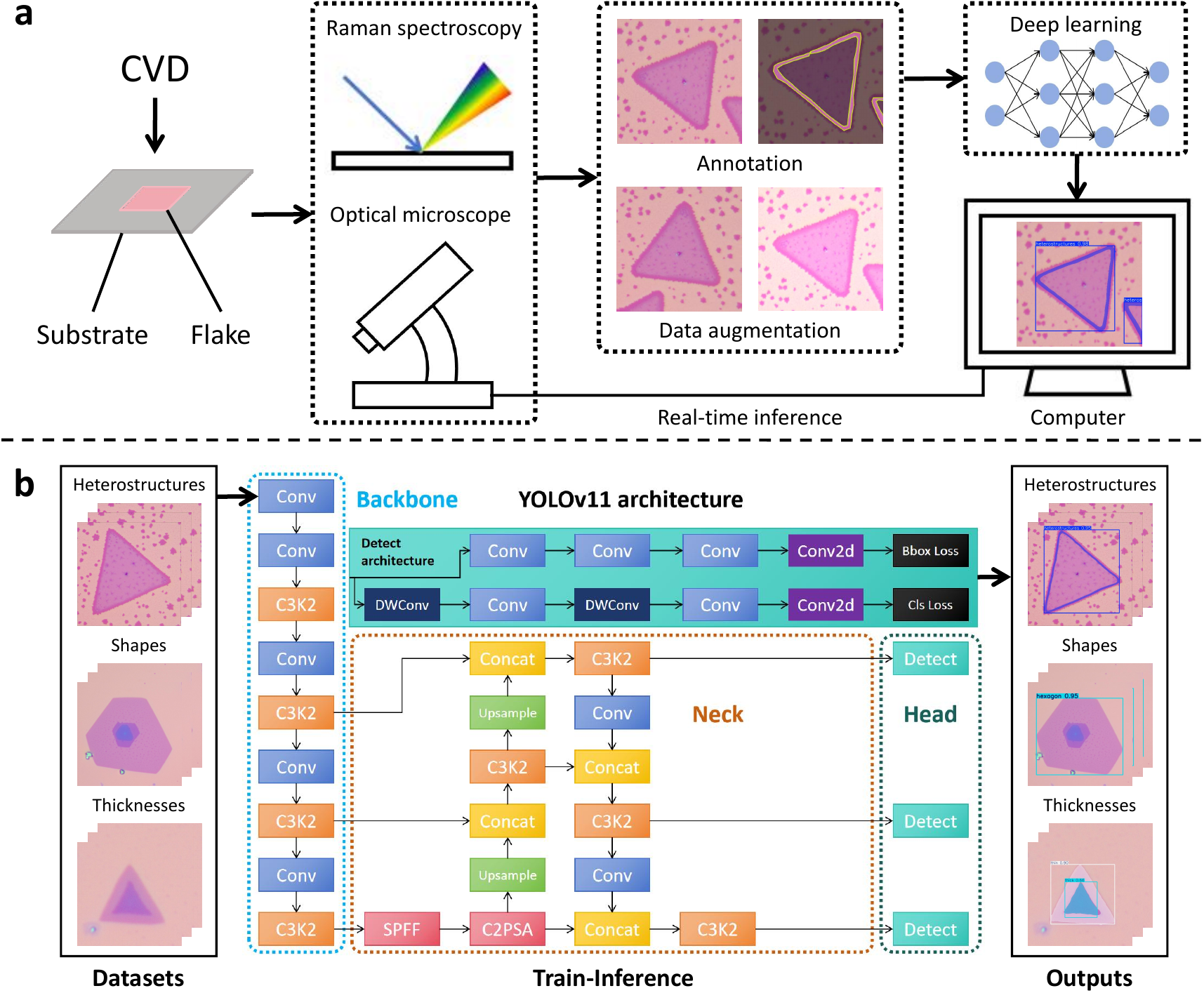}
    \caption{Experimental workflow and deep learning architecture.(a) Key steps: Material preparation $\rightarrow$ structural characterization $\rightarrow$ dataset annotation $\rightarrow$ YOLOv11 training $\rightarrow$ real-time deployment on optical microscopy systems.  (b) YOLOv11 architecture schematic showing data input (left), model architecture (center), and inference results (right).}
    \label{Fig1}
\end{figure*}

The experimental process is depicted in Figure \ref{Fig1}a. Initially, an optical microscope was utilized to capture a large number of optical images of 2D materials. Lateral heterostructures were characterized using Raman spectroscopy, and the features were annotated based on the characterization results to train them in instance segmentation tasks. The optical images of the shapes were classified into three categories for annotation: dendritic, hexagonal, and triangle. Once the annotation was completed, the object detection task was proceeded with. Transfer learning was applied to the dataset of material thicknesses, which were classified as thin, thick, and bulk, followed by training for the instance segmentation task. The training process and model architecture are illustrated in Figure \ref{Fig1}b. Considering the hardware environment and dataset sizes, only the three smallest YOLO models, namely n, s, and m, were used for training. The difference between them lies in the size of the neural network, with the network becoming progressively deeper and wider from n to m. The models architectures are shown in Supplementary Information Section 1 and 2.

\subsection{Training results}

\begin{figure*}[!thp]
    \centering
    \includegraphics[width=0.75\textwidth]{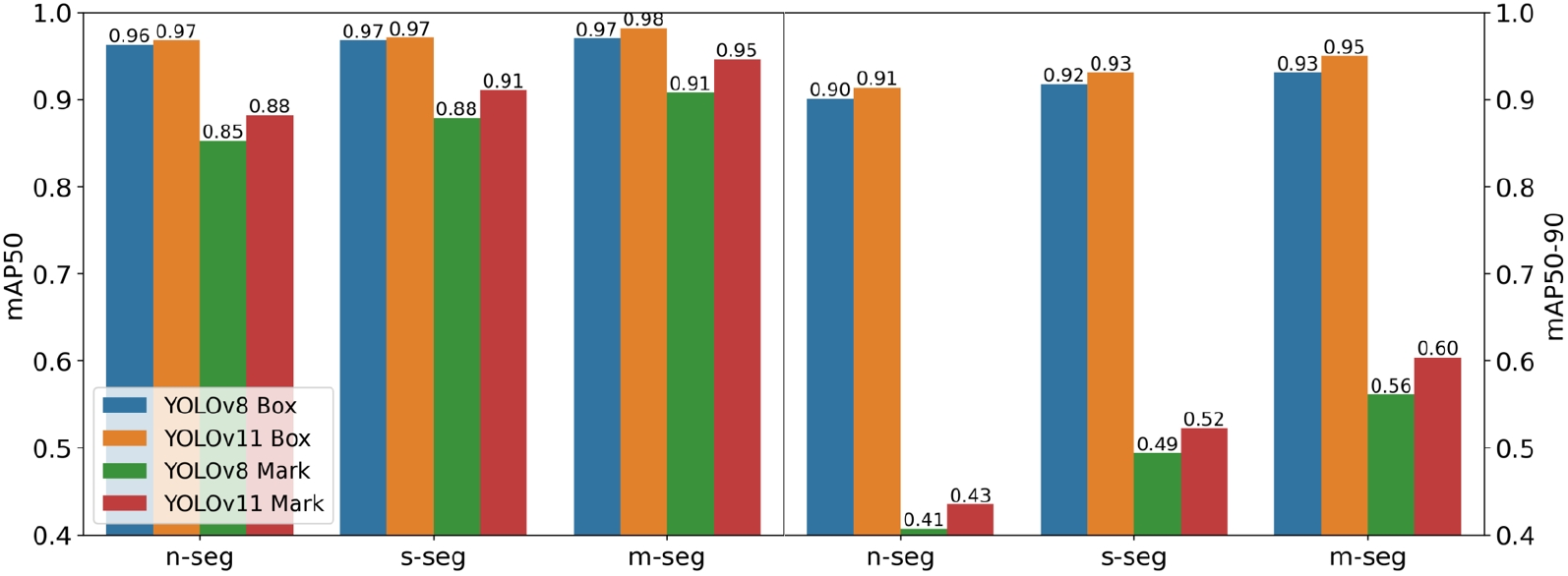}
    \caption{Performance comparison of heterostructure characterization models. Among them, box refers to boundary box detection, and mark refers to mask segmentation. The left image shows mAP50, and the right image shows mAP50-90.}
    \label{Fig2}
\end{figure*}

The training results for the MoS$_2$-MoSe$_2$ heterostructures dataset are shown in Figure \ref{Fig2}. It is evident that the performance indexes of all models are at a relatively high level. Among them, mAP50 and mAP50-90 are important indicators for evaluating object detection algorithms, representing the overlap between predicted boxes and real boxes. The specific calculation process is shown in the Supplementary Information Section 3. Under the same model size, YOLOv11 outperforms YOLOv8 significantly, with an average improvement of 3.74\% in boundary box detection and 6.80\% in mask segmentation for mAP. From the perspective of model size analysis, the average improvement in mAP50 for mask segmentation is 3.37\% as the model size increases, while the average improvement in mAP50-90 is 17.46\% with larger model sizes. Among these, YOLOv8s-seg shows a 21.37\% improvement compared to YOLOv8n-seg. This suggests that increasing the model's complexity can effectively enhance its ability to infer and identify heterostructures. Of all the models, YOLOv11m-seg achieved the best performance. From the confusion matrices (Figure S3), it can be observed that the model's recognition accuracy exceeds 95.6\%, indicating excellent performance. 

\begin{figure}[!thp]
    \centering
    \includegraphics[width=0.65\textwidth]{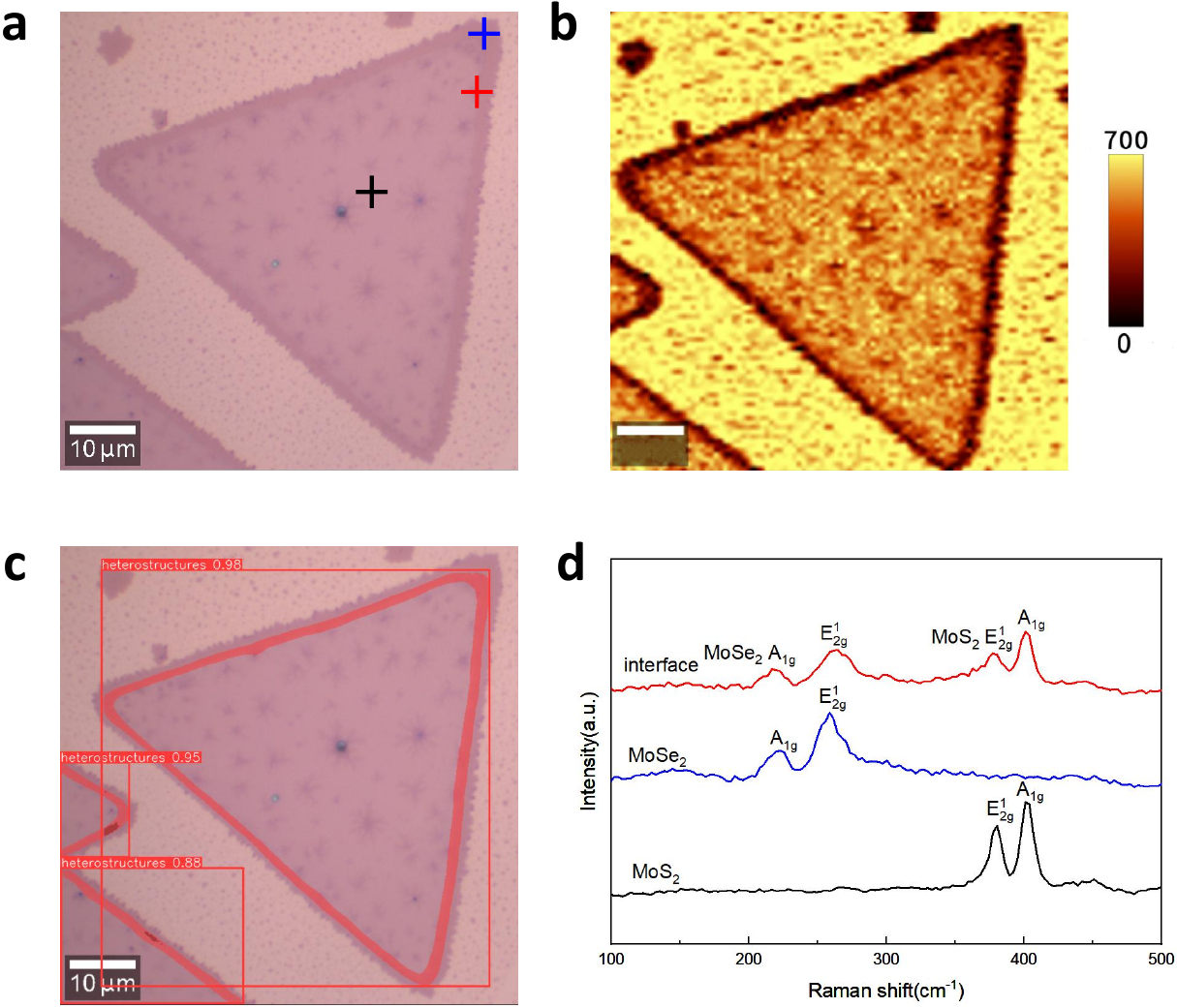}
    \caption{Characterization of lateral heterostructures of MoS$_2$-MoSe$_2$ and YOLO model inference results. (a) Optical images of the heterostructures. (b) Raman mapping of the heterostructures. (c) Heterostructures predicted based on the YOLO model. (d) Raman spectra, where the measurement areas are marked in red, blue, and black in (a).}
    \label{Fig3}
\end{figure}

Figure \ref{Fig3} shows the actual inference results of the YOLO model. The morphology of the lateral heterostructures of MoS$_2$-MoSe$_2$ was characterized by optical microscopy and Raman spectroscopy. Figure \ref{Fig3}a and b present the optical images and corresponding Raman mapping images of the heterostructures, respectively. Figure \ref{Fig3}c shows the inference results. The Raman spectra measured at the center, periphery, and boundary areas of the triangle is shown in Figure \ref{Fig3}d. The ${A}_{1g}$ mode (out-of-plane vibration) and ${E}_{2g}^{1}$ mode (in-plane vibration) of MoS$_2$ appear in the center area, while the ${A}_{1g}$ mode and ${E}_{2g}^{1}$ mode of MoSe$_2$ appear in the periphery area. At the junction, four peaks corresponding to the ${A}_{1g}$ and ${E}_{2g}^{1}$ modes of MoS$_2$, as well as the ${A}_{1g}$ and ${E}_{2g}^{1}$ modes of MoSe$_2$, confirm the presence of a lateral heterostructure\cite{chen2019lateral}. The Raman mapping characterization in Figure \ref{Fig3}b further confirms the coexistence of internal MoS$_2$ and external MoSe$_2$, with clear visibility at the interface, revealing the spatial distribution of heterostructures and indicating the seamless lateral coexistence of two different materials within the same triangular domain.

\begin{figure*}[!thp]
    \centering
    \includegraphics[width=0.75\textwidth]{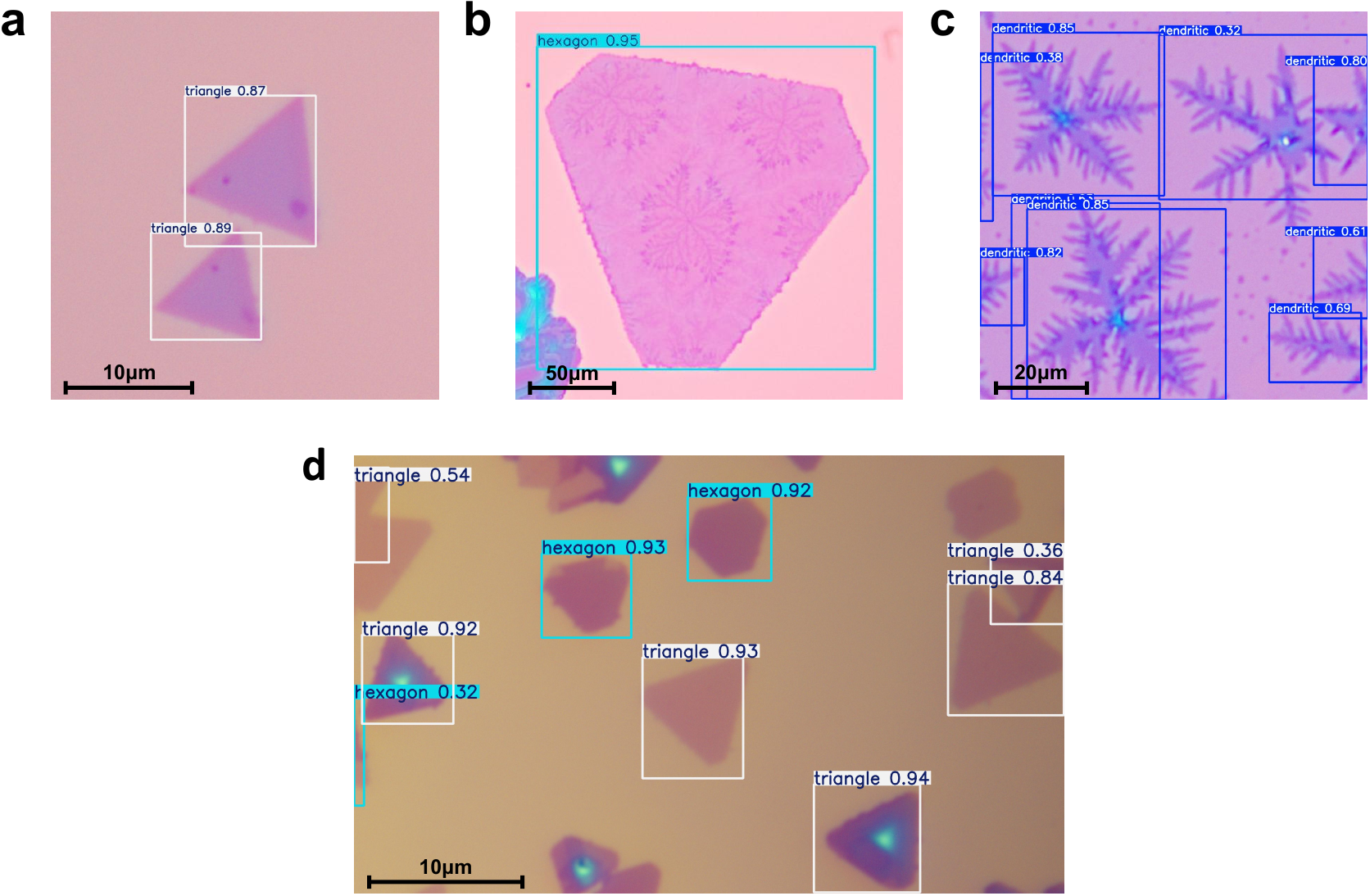}
    \caption{Identification results of MoS$_2$ with different shapes: (a) triangle, (b) hexagon, (c) dendritic. (d) Optical image inference results of MoS$_2$.}
    \label{Fig4}
\end{figure*}

The morphology and spatial distribution of MoS$_2$ nanostructures are highly sensitive to synthesis parameters such as growth duration, temperature gradients, precursor stoichiometry, and the positioning of the substrate within the reaction zone \cite{wang2014shape,cao2015role,wang2017shape,cho2023size}. Recent advancements in deep learning-based shape identification techniques facilitate the rapid mapping of synthesis conditions to nanostructure geometries, providing real-time feedback for process optimization.
The results of training the shape dataset using YOLOV11 are shown in Table S4-S5. The confusion matrix for YOLOv11s (Figure S4) shows an average recognition accuracy of 94.67\%. Figures \ref{Fig4}a-c display the model's ability to effectively identify MoS$_2$ shapes with high confidence. By analyzing the recognized boundary boxes, the area proportions of each shape can be calculated. For example, as shown in Figure \ref{Fig4}d, triangles account for 17.60\% of the image area, while hexagons cover 4.86\%. This rapid identification and statistical method based on deep learning can aid in regulating the growth of MoS$_2$ or other 2D materials with different shapes at high temperatures\cite{moses2024quantitative}.

\begin{figure*}[!hbtp]
    \centering
    \includegraphics[width=0.75\textwidth]{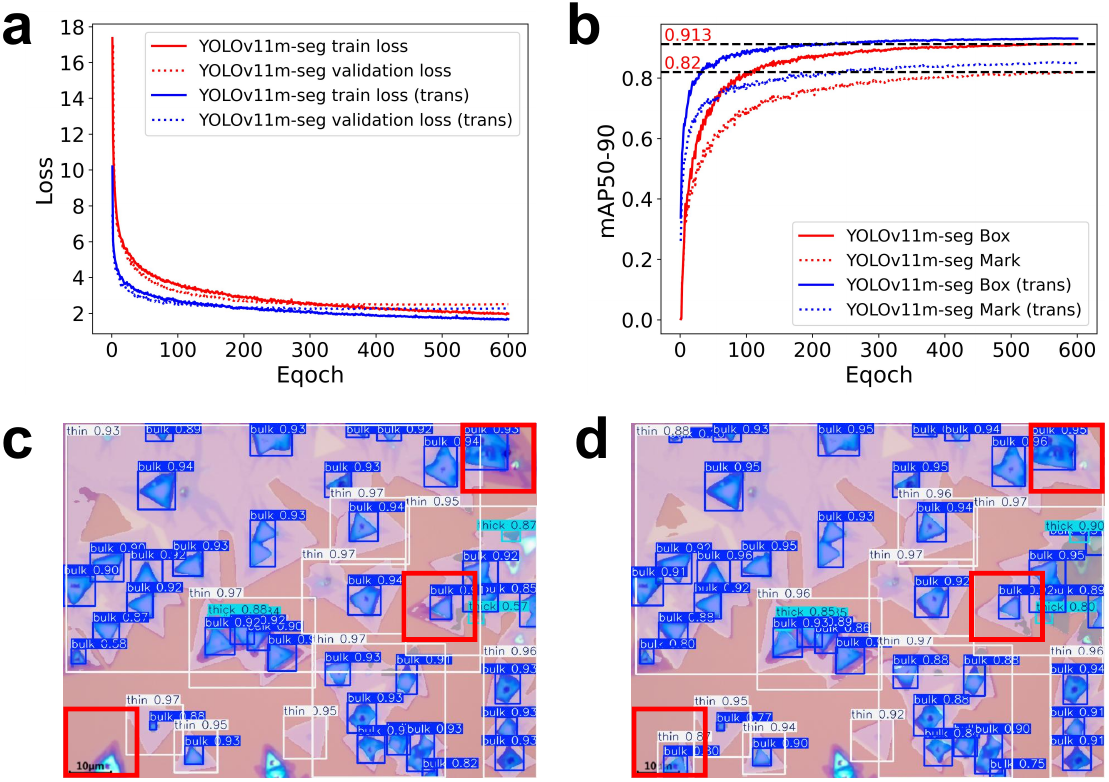}
    \caption{Training results of the MoS$_2$ thicknesses dataset. (a) Loss curves, where the solid line indicates the training set loss and the dashed line represents the validation set loss. (b) mAP50-90 curves, with the solid line showing the mAP50-90 for the boundary box detection and the dashed line for mask segmentation. In (a) and (b), the red line represents the original model, and blue line denotes the transfer learning model. (c) YOLOv11m-seg model inference results. (d) Inference results of YOLOv11m-seg model after transfer learning. Differences between (c) and (d) are highlighted with red boxes.}
    \label{Fig5}
\end{figure*}

The YOLOv11 models trained with the heterostructures dataset were used for transfer learning on the thicknesses dataset, with results shown in Table S7. Figure \ref{Fig5}a compares the loss curves for direct training and transfer learning with the YOLOv11m-seg model. The transfer learning model shows faster loss reduction, outperforming the original model in the initial phases. After 600 epochs, the loss decreased by 17.84\% for the training set and 11.38\% for the validation set compared to the original model. Additionally, the transfer learning model excels in performance, as shown in Figure \ref{Fig5}b, with better mAP50-90 results. After the 189th and 228th epochs, it surpassed the highest mAP50-90 achieved by the original model in both boundary box detection and mask segmentation. After 600 epochs, the transfer learning model led by 2.08\% in detection boundary box mAP50-90 and 3.90\% in mask segmentation mAP50-90. Inference results (Figures \ref{Fig5}c-d) for both models were almost identical, but the transfer learning model identified more instances, detecting 3 more bulk instances and 1 more thin instance. The confusion matrix (Figure S5) shows an average accuracy of over 97.3\% for both models, but transfer learning achieved this in less than half the training time. Transfer learning was also conducted on the WTe$_2$ thickness dataset (Figure S6), where it did not significantly improve performance compared to the original model. However, in early training, the loss decreased faster, and mAP50-90 improved more quickly, indicating that transfer learning is less effective across different materials. Overall, transfer learning enhances model performance within the same material, but its impact is less significant when applied between different materials.

\subsection{Generalization ability}

\begin{figure*}[!thp]
    \centering
    \includegraphics[width=0.8\textwidth]{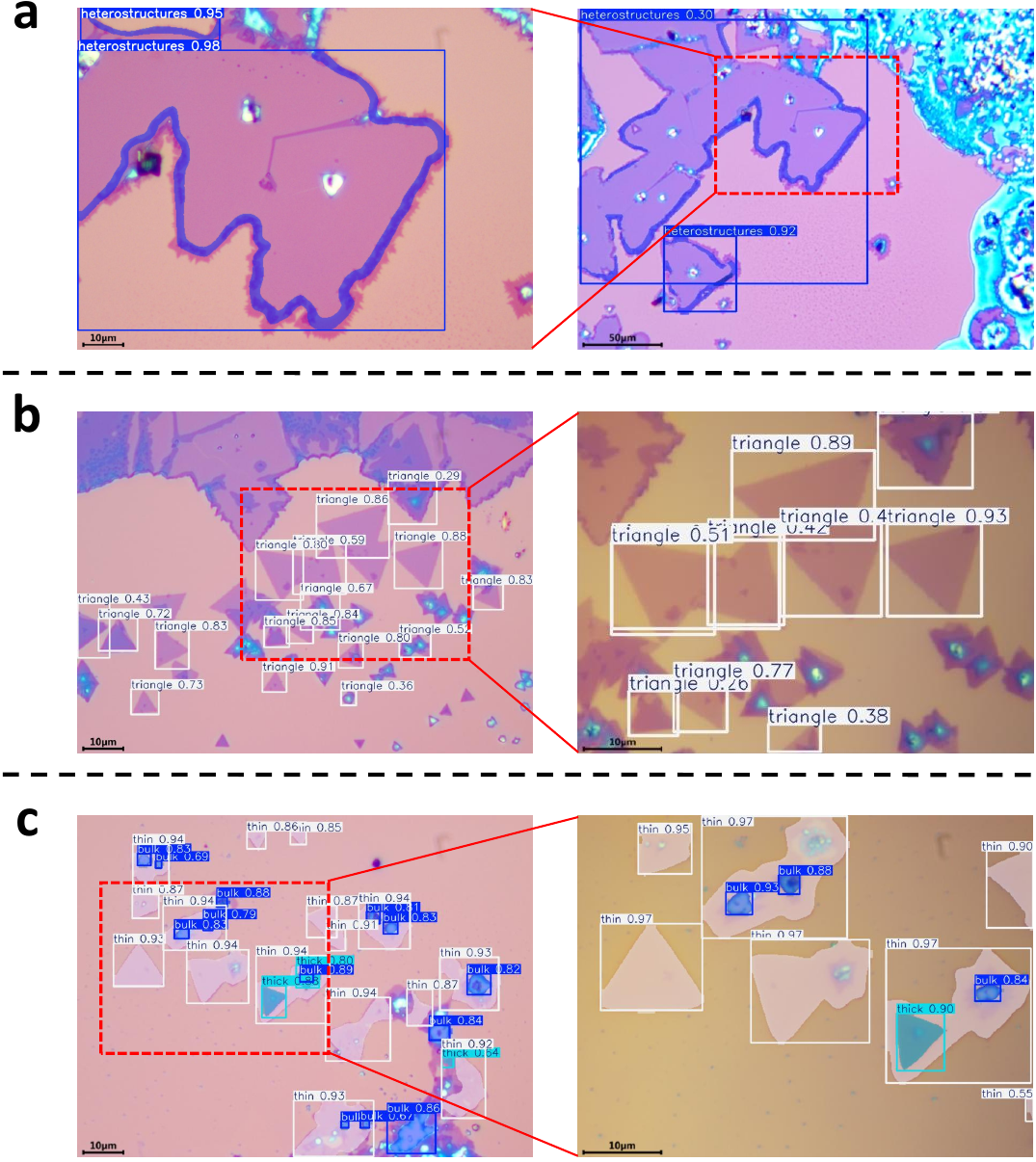}
    \caption{YOLO model's generalization ability in various environments: (a)-(c) show the generalization ability in large-scale complex images. The first column (500$\times$ magnification) shows heterostructures, shapes, and thicknesses (top to bottom). The second column displays results at different magnifications: 200$\times$, 1000$\times$, and 1000$\times$ (top to bottom). The red box indicates the zoomed-in area.}
    \label{Fig6}
\end{figure*}

Figure \ref{Fig6} demonstrates the generalization ability of the model across different scenarios. Among them, Figure \ref{Fig6}a-c show the generalization ability of the YOLO models in complex scenes with heterostructures, shapes, and thicknesses. It is evident that the YOLO models exhibit excellent generalization, successfully recognizing various targets in complex environments. The second column of Figure \ref{Fig6} displays inference results for optical images at various magnifications. At different magnifications, the propagation path of light in the microscope varies, leading to light scattering and chromatic aberration. However, the model still maintains strong performance at magnifications of 200$\times$, 1000$\times$, and 1000$\times$, demonstrating effective inference under different lighting conditions.

\begin{figure*}[!thb]
    \centering
    \includegraphics[width=0.9\textwidth]{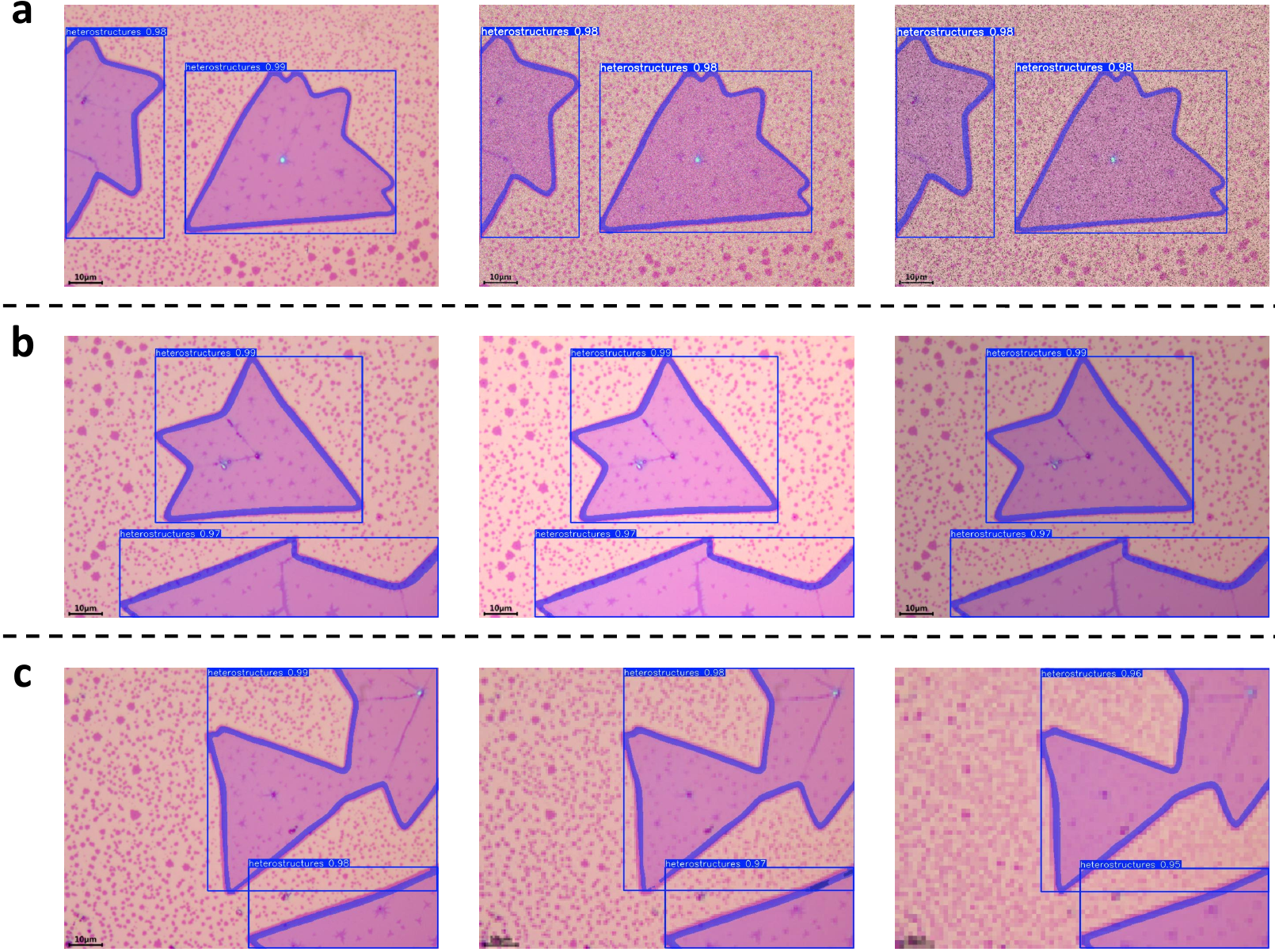}
    \caption{Generalization ability of the YOLO model on anomalous optical images of heterostructures. (a) Robustness to noise: the first image shows the original inference result, the second with 20\% Gaussian noise, and the third with 20\% salt-and-pepper noise (50\% salt, 50\% pepper). (b) Sensitivity to brightness: the first image represents the original inference, the second with a 20\% brightness increase, and the third with a 20\% decrease. (c) Performance at different resolutions: the first image is the original inference, while the second and third correspond to resolution reductions to 1/5 and 1/10 of the original, respectively.} 
    \label{Fig7}
\end{figure*}

Figure \ref{Fig7} shows the generalization ability of the models under three interference factors: noise, brightness, and resolution. It can be observed that the models' inference results under interference are almost identical to the original image, demonstrating strong anti-interference ability. We conducted separate tests on the characterization of shape and thickness (Figure S7-S8), and the model also demonstrated excellent performance. This indicates all models have good robustness and anti-interference ability.

\subsection{Application}

Optical microscopes can be integrated with deep learning models for real-time inference. we propose a novel method based on YOLO models for real-time inference, implemented using Python. This approach not only achieves high-frame-rate speeds but also supports local deployment, eliminating the time and resource costs of image transmission.

\begin{figure*}[!thp]
    \centering
    \includegraphics[width=0.9\textwidth]{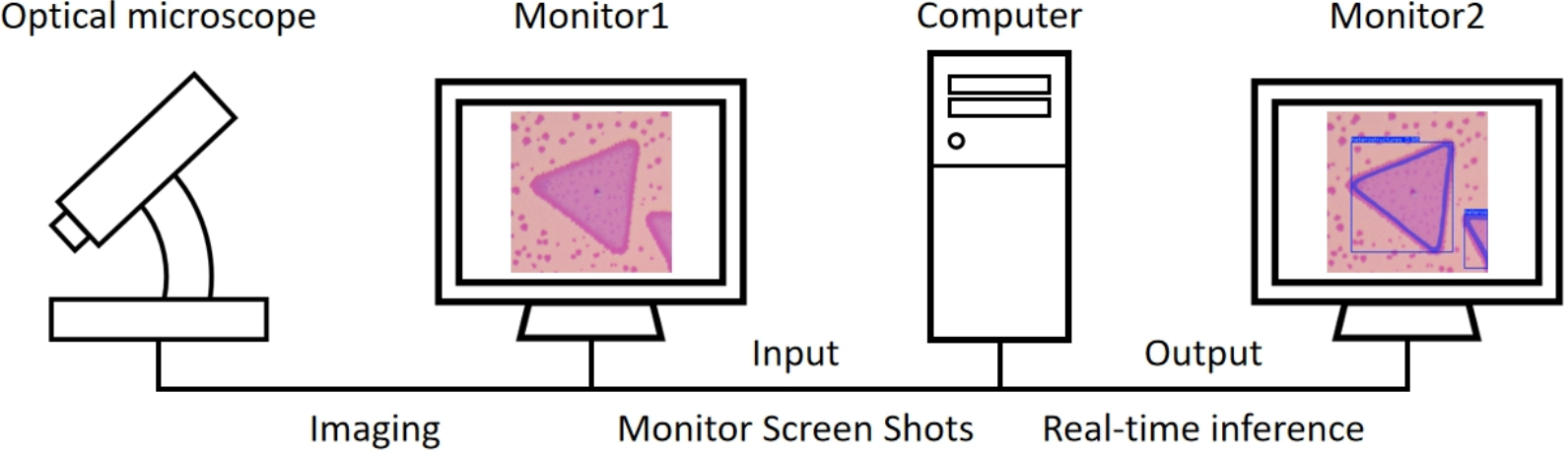}
    \caption{Real-time inference process. Optical images are captured using an optical microscope and transferred to a computer (displayed on Monitor 1). The MSS library then screenshots these images and feeds them into the YOLO model for inference. The results are shown on Monitor 2. By continuously capturing screenshots and running inferences multiple times per second, a real-time video stream is generated.}
    \label{Fig8}
\end{figure*}

The real-time inference process is shown in Figure \ref{Fig8}. It involves capturing the microscope screen using the MSS (Monitor Screen Shots) library, feeding the screenshot into the YOLO model, and outputting the processed image in real-time via OpenCV. The experimental setup is shown in Figure S10, and Supplementary Video 1 demonstrates the real-time inference functionality. Thanks to the efficient YOLO models, we can deploy deep learning locally for real-time processing. Table S10 shows the inference speed and FPS on different computers, with the low-complexity model achieving up to 50 FPS.

\section{Conclusions}

In summary, this article proposes a novel deep learning-based method for rapidly characterizing MoS$_2$-MoSe$_2$ lateral heterostructures. We also use deep learning to characterize the shape and thickness of MoS$_2$ flakes, and find that transfer learning can effectively enhance accuracy. The experimental results show that the accuracy of heterostructures characterization method reaches 95.6\%, with shape recognition accuracy of 94.67\% and thickness recognition accuracy of 97.3\%. After testing, the trained models demonstrate strong robustness and resistance to interference. Additionally, for convenience in characterization, we integrated optical microscopy with deep learning models to achieve real-time inference. This characterization method based on deep learning provides a faster and more convenient way to characterize the heterostructures and other physical properties of 2D materials.

\section{Methods}

\subsection{Preparation and characterization}

The MoS$_2$ flakes and MoS$_2$-MoSe$_2$ lateral heterostructures were synthesized via atmospheric pressure chemical vapor deposition (APCVD) in a dual-temperature zone horizontal tube furnace. MoO$_3$ powder (Alfa Aesar, 99.998\%) mixed with NaCl (Alfa Aesar, 99.99\%) was placed in an alumina boat in the high-temperature zone with a SiO$_2$/Si substrate positioned above it\cite{cai2018synthesis}. Sulfur (Alfa Aesar, 99.999\%)(240 $^\circ$C zone) and selenium powders (Alfa Aesar, 99.999\%) were placed in separate upstream alumina boats, with sulfur closer to MoO$_3$\cite{wang2017nacl}. The furnace was heated at 25 $^\circ$C/min under a pure argon flow 120 (standard cubic centimeters per minute (sccm)). After 15 minutes of MoS$_2$ growth, the atmosphere was switched to N$_2$/H$_2$ (9:1) gas mixture for 10 minutes to terminate MoS$_2$ formation\cite{duan2014lateral}. The Se zone temperature was then raised to 350 $^\circ$C (substrate zone maintained at 650 $^\circ$C ) for 15 minutes for MoSe$_2$ edge-selective growth under Se vapor. Finally, natural cooling yielded the heterostructures\cite{chen2017plane,liu2023controllable}.

The optical images were acquired using an optical microscope (AOSVI, L100-3M100). Raman spectroscopy was performed using a WITEC alpha 300R micro Raman spectrometer equipped with a 532 nm laser (10 mW power) at room temperature.

\subsection{Dataset preprocessing}

We selected 180 heterostructure images with distinct features, 50 MoS$_2$ optical images with different shapes, and 50 MoS$_2$ optical images with different thickness. Split all images into squares of size 640$\times$640. Use the Roboflow online annotation tool for annotation, and save the boundary box coordinates and masks of each instance in YOLO format. Then perform data augmentation on the dataset, as shown in Figure S12, which was expanded to 3955, 1101, and 1103 images, respectively. We divided the datasets into a training set and a validation set in an 8:2 ratio, and used the remaining unselected images as the test sets.

\subsection{Training environment}

Software environment with Python 3.9, PyTorch 2.5.0, and CUDA 11.7. Hardware environment with NVIDIA GeForce RTX 3060 GPU, 64GB memory. The hyperparameters are shown in Table S1, S3, and S6. Real-time inference programs based on Python rely on libraries such as PyQt5, OpenCV, MSS, and PyTorch for implementation. In Supplementary Video 2, all the functions and usage methods of the application are demonstrated.

\begin{acknowledgement}
Xiaohui Yang acknowledges the support from the Zhejiang Provincial Natural Science Foundation of China (Grant No. LQ23A040009) 
and the National Natural Science Foundation of China (Grant No. 12304168). Jialu Wang acknowledges the support from the National Natural Science Foundation of China (Grant No. 12404047).

\end{acknowledgement}

\section*{Data Availability}
The datasets used in this study are available from the corresponding author upon reasonable request. The codes underlying this study are openly available in Github at \url{https://github.com/TripHawkers/OM-real-time-inference}.

\begin{suppinfo}

\begin{itemize}
  \item Supplementary Information: Deep learning algorithms, Optimizer, Performance index, Model training results, Generalization ability, Real-time inference, Data augmentation.
  \item Video 1: real-time inference
  \item Video 2: application function demonstration
\end{itemize}

\end{suppinfo}

\section{Author Contributions}

J.H. and Y.Z. contributed equally to this work. J.H. and X.Y. conceived the project; Y.Z. synthesized the samples and provided optical images with the help of C.C. and J.Y.; Y.Z. performed the Raman measurements with the help J.W. and T.W.; J.H. implemented the software, trained the neural network, and analyzed the results with the help from P.Z.; X.Y. supervised the research program. All authors participated in the discussion of the results and the writing of the manuscript.

\section{Competing interests}
The authors declare no competing interests.


\providecommand{\latin}[1]{#1}
\makeatletter
\providecommand{\doi}
  {\begingroup\let\do\@makeother\dospecials
  \catcode`\{=1 \catcode`\}=2 \doi@aux}
\providecommand{\doi@aux}[1]{\endgroup\texttt{#1}}
\makeatother
\providecommand*\mcitethebibliography{\thebibliography}
\csname @ifundefined\endcsname{endmcitethebibliography}  {\let\endmcitethebibliography\endthebibliography}{}

\end{document}